\documentclass[journal=jctcce,manuscript=article]{achemso}

\usepackage[T1]{fontenc} 
\usepackage{hyperref}       
\usepackage{graphicx}       
\usepackage{url}            
\usepackage{booktabs}       
\usepackage{amsfonts}       
\usepackage{natbib}
\usepackage{subcaption}
\usepackage{standalone}
\usepackage{amsmath}
\usepackage{siunitx}
\usepackage{import}
\usepackage[draft]{minted}
\usepackage{multirow}
\usepackage{wrapfig}
\usepackage{siunitx}
\usepackage{amssymb}
\usepackage{caption}
\usepackage{graphicx} 

\author{Zehua Zhang}
\affiliation[INI]{Information Network Institute,
Carnegie Mellon University,
Pittsburgh PA, USA}

\author{Zijie Li}
\affiliation[MechE]{Department of Mechanical Engineering,
Carnegie Mellon University,
Pittsburgh PA, USA}

\author{Amir Barati Farimani}
\affiliation[MechE]{Department of Mechanical Engineering,
Carnegie Mellon University,
Pittsburgh PA, USA}

\alsoaffiliation[ML]{Machine Learning Department,
Carnegie Mellon University,
Pittsburgh PA, USA}
\alsoaffiliation[Chem]{Department of Chemical Engineering,
Carnegie Mellon University,
Pittsburgh PA, USA}
\email{barati@cmu.edu}
\title
  {Masked Pretraining Strategy for Neural Potentials}

\begin{document}

\begin{abstract}
We propose a masked pretraining method for Graph Neural Networks (GNNs) to improve their performance on fitting potential energy surfaces, particularly in water and small organic molecule systems. GNNs are pretrained by recovering spatial information of masked-out atoms from molecules selected with certain ratios, then transferred and finetuned on atomic forcefields. Through such pretraining, GNNs learn meaningful prior about structural and underlying physical information of molecule systems that are useful for downstream tasks. With comprehensive experiments and ablation studies, we show that the proposed method improves both the accuracy and convergence speed of GNNs compared to their counterparts trained from scratch or with other pretraining techniques. In addition, our pretraining method is suitable for both energy-centric and force-centric GNNs. This approach showcases its potential to enhance the performance and data efficiency of GNNs in fitting molecular force fields.
\end{abstract}

\section{Introduction} 
\
Molecular dynamics plays a pivotal role in elucidating the dynamic behavior of molecules, providing essential insights into the temporal evolution of complex systems, such as understanding the conformational changes, interactions, and thermodynamic properties of molecules. It is thus important for fields ranging from drug discovery to materials science. The potential energy surface serves as the underlying landscape dictating the dynamics of a molecular system, and making an accurate representation of it is crucial for studying the dynamics of molecules through numerical simulations. Ab initio methods like Density Functional Theory (DFT) provide high accuracy by accounting for the electronic structure of atoms. Yet their high computational demands pose a significant challenge for efficient utilization, particularly in the context of large many-body systems. On the other hand, forces can be computed using empirical interatomic potentials tailored to specific environments, bypassing the electronic structures of the system. This approach significantly reduces the computational cost compared to ab initio methods. As there is a wide variety of interactions in molecular systems (bonded or non-bonded interactions), finding the appropriate functional forms can be challenging \cite{review-FFs-2018, SpookyNet-2021}. The difficulty in describing complex interatomic interaction presents a potential limitation for classical MD. Machine learning (ML) has emerged as a promising alternative for learning and fitting the potential energy surface with accuracy close to ab initio approaches \cite{review-Noe-2020, review-Barati-2020, unke2021-ML-Forcefield, ML-Forcefield-2017-Botu, ML-forcefield-Ab-initio-2017, ML-forcefield-nature-2018, ML-interatomic-2019, ml-potential-review, nnp-sodium, nnp-zinc, DeepPMD, ManySpecies-mlp}. The kernel-based regression model is one of the examples. Gradient-domain ML (GDML) employs a kernel-based method to fit the forcefield with the guarantee of energy conservation \cite{Chmiela-GDML-2017}. Following GDML, Symmetric GDML (sGDML) extends GDML to incorporate space group symmetries and dynamic nonrigid symmetries \cite{chmiela2018towards}. Gaussian approximation potentials (GAPs) decompose the energy as the sum of atomic-centered Gaussian basis function \cite{bartok2010gaussian, bartok2017machine} with hand-designed local environment descriptors \cite{bartok2013representing, grisafi2018symmetry}. 
The strong fitting capacity of the neural network has made it a suitable choice for fitting the potential energy surface. The initial efforts of building neural-network-based potential (neural potential) have leveraged tailored descriptors that capture the characteristics of local environment of atoms in the system \cite{Bartok-ML-modeling, ML-forcefield-Ab-initio-2017, ANI-1-Smith-2017, FCHL-revisited-2020, DPMD-2018, fingerprints-o2}, where Behler-Parrinello Neural Networks (BPNN) \cite{Behler-2011-ACSF, Behler-NN-Energy-2007, Behler-BPNN2017} is one of the first examples. The recent advancements of deep neural networks, especially graph neural networks (GNNs), open up the possibility of learning the atomic representations directly from the raw atomic coordinates, which provides an alternative to the manually tailored atomic fingerprints \cite{Conv_fingerprints_NIPS-2015, Molecular-graph-2016, Schutt-insight-DTNN-nature, PhysNet-2019-Unke, schutt2017schnet, Hierarchical-2018-JCP, atomic-fingerprints-2018-ML}. In GNNs, the interaction between atoms is modeled by message passing \citep{Gilmer-MP-ICML-2017} between atoms within the local neighborhood. SchNet \citep{schutt2017schnet} introduces continuous convolution for aggregating messages across nodes (atoms) to produce a continuous energy surface. Various other works have developed different message-passing protocols based on physics and chemical insights \citep{Schutt-insight-DTNN-nature, PhysNet-2019-Unke, Hierarchical-2018-JCP, atomic-fingerprints-2018-ML, Directional-MPNN-ICLR-2020}. One notable variant is the class of equivariant GNNs. With tailored message-passing functions and update strategies, these neural networks are equivariant to group actions such as rotation and translation in 3D Euclidean groups. The exploitation of property of the symmetry has made equivariant GNNs not only more data-efficient and also more accurate \citep{thomas2018tensor, fuchs2020se, anderson2019cormorant, jing2021learning, villar2021scalars, gasteiger2021gemnet, batzner20223, cheng2024cartesian, batatia2022mace}. 

While GNN-based neural potentials have provided a flexible and accurate end-to-end framework for learning and predicting the potential energy, their performance heavily relies on the amount of data. Collecting DFT-based data can be expensive or even infeasible for large and complex molecular systems. Pretraining and finetuning have been proven to be an effective popular paradigm that can facilitate better data efficiency \cite{zhang2023dpa2, zhang2023dpa1, feng2023force}. To specifically improve the data efficiency of GNNs on molecular modeling tasks such as property prediction, various pretraining strategies have been investigated. One direction is pretraining GNNs by predicting the masked attributes or context of graphs \citep{Hu2020Strategies, rong2020self, zhang2021motif}. Another direction is pretraining GNNs by contrasting the embeddings from the network given different views of the data \citep{wang2021molclr, liu2022graph, krishnan2022self, magar2022crystal, cao2022moformer, zhang2020motif, wang2022improving, zhang2021motif}. Different views can be generated via randomly masking out some of the nodes and edges in the molecular graphs \citep{wang2021molclr} or directly contrasting 3D representation with 2D representation of the molecular system \citep{stark20223d, liu2022pretraining}. It is worth noting that many of these methods have focused on improving the GNNs' performance on 2D graphs, while for neural potential the input is usually 3D structures (coordinates of atoms) of molecular systems. Several recent works have proposed to pretrain the GNNs on 3D representation by denoising \citep{zaidi2022pre, liu2022molecular, zhou2023unimol, Wang2023denoise}. More specifically, the pretraining task is to predict the atom-wise Gaussian noise added to the atom coordinates under the equilibrium state or non-equilibrium state. The denoising pretraining has been demonstrated to help improve models' performance in predicting the properties of small molecules or single molecule systems. Previous works have viewed denoising as learning the pseudo-gradient (or force field) of the energy surface, as an appropriately small perturbation (e.g. Gaussian noise with variance less than 0.1) can yield meaningful learning tasks for GNNs. However, identifying the appropriate noise scale can be difficult, such that it should not be trivially small nor excessively large. For the water system, the overall potential energy includes the contribution from intra-molecule interaction and inter-molecule interaction. This makes selecting an appropriate variance much trickier than single-molecule systems, given that small noise cannot yield significant enough perturbation to the intra-molecule part and large perturbation is too difficult for GNNs to recover the original geometric structures.

In this work, we propose to pretrain GNNs with tasks of recovering the spatial information of selectively masked-out atoms, and we showcase that it's robust and versatile for different molecular systems, from water systems to organic molecules. In addition, we observe that our proposed pretraining strategy yields consistent improvement on both non-equivariant and equivariant GNNs.

\begin{figure}[h]
    \centering
    \includegraphics[width=1\linewidth]{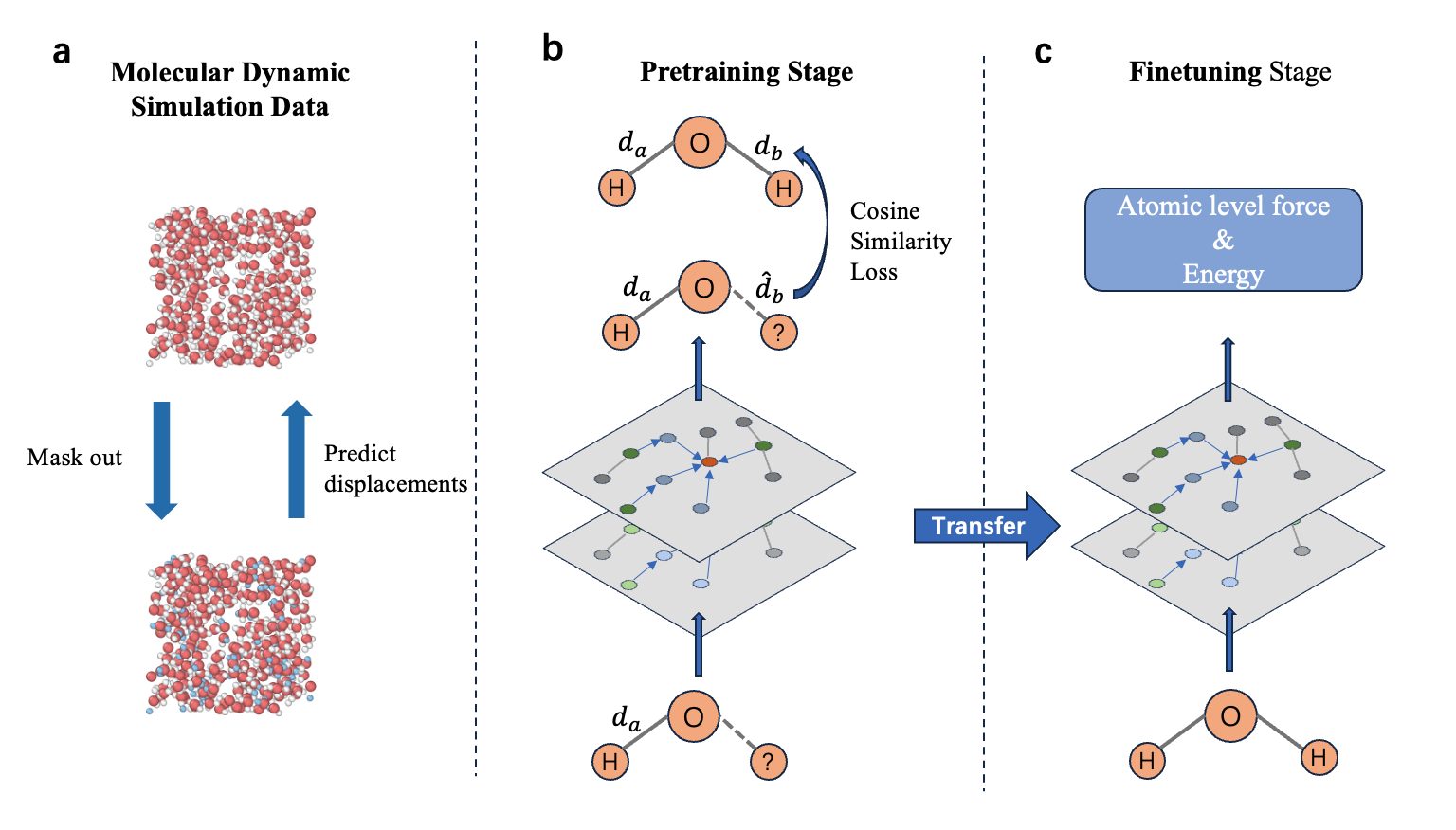}
    \caption{Framework of masked pretraining GNNs on water molecule systems. (a) With water molecules selected on a certain ratio, spatial information of one Hydrogen atom (Blue) in each selected water molecule is masked to create a pretext task. (b) GNNs are pretrained to recover displacements between masked-out atoms and other atoms. Information of any atoms but the masked-out ones is provided for GNNs to predict such displacements. (c) The pretrained weights are transferred and finetuned to predict the potential energy surface. } 
    \label{fig:masked architecture}
\end{figure}

\section{Methodology}

\subsection{Graph Neural Networks}
Molecules are naturally suitable to be modeled as graphs, where atoms are depicted as nodes and their interactions as edges. Message Passing Neural Network (MPNN)\citep{Gilmer-MP-ICML-2017}, has been a popular Graph Neural Network paradigm for its competitive performance on complexly structured graphs, which makes it apt for molecular property prediction. It often includes two phases during forward propagation: message passing and readout. We define the constructed graph \(G\) that contains node features \(x_v\) and edge features \(e_{vw}\) encoded from atoms and bonds (interaction) between atoms. During the message passing stage, hidden states \(h_v^t\) at each node are updated with an update function \(f_u^t\) based on messages \(m_v^{t+1}\) that are aggregated by a message function \(f_m^t\). Formally:

\begin{equation}
    m_{v}^{(t+1)} = \sum_{j \in \mathcal{N}(i)} f_{m}^{(t)}(h_{v}^{(t)}, h_{w}^{(t)}, e_{vw}) 
\end{equation}

\begin{equation}
    h_{v}^{(t+1)} = f_{u}^{(t)} \left( h_{v}^{(t)}, m_{v}^{(t+1)} \right) 
\end{equation}

Then a readout function \(R\) aggregates all node features to learn a graph-level representation:
\begin{equation}
    \hat{y} = R(\{h_v^T | v \in G\}) 
\end{equation}

Several research \citep{Dejun2021molecularRep,rong2020selfsupervised,Hu2020Strategies} focused on extending MPNNs to the field of molecular attribute predictions. Among them, a subset of equivariant and covariant MPNNs such as EGNN\citep{satorras2021n}, GNS\citep{kumar2022gns}, and ForceNet\citep{hu-2021-forcenet} are used in our investigation as baseline models. EGNN is an energy-centric model that extends equivariance to higher orders and shows competitive performance on molecular datasets such as QM9 \citep{Rama2014QM9}. When used to predict molecular dynamics, EGNN first predicts energy while maintaining rotational equivariant and energy conservation. The forces are then derived as the negative first derivatives of energy with respect to atomic positions. On the other hand, both GNS and ForceNet are based on encode-process-decode architecture and are force-centric. Specifically, a GNN is used to encode and update node embeddings, which are then decoded and used to predict atomic forces; a Multilayer Perceptron (MLP) finally approximates the energy by aggregating decoded embeddings. 

\subsection{Pretraining by masking}  \label{Sec: Masking Methodology}
Our proposed approach Hydrogen atom masking, similar to token masking in Masked Language Modeling \citep{devlin2019bert}, refers to deliberately masking one Hydrogen atom out of each selected water molecule with its spatial information removed from the input. The selection ratio is a hyperparameter that directly affects our framework's performance. Analysis of the choice masking ratios is further discussed in the supplementary information Section 1.

Specifically, before masking, we calculate displacements $\mathbf{d}_i$ between atoms within the same molecule and append them to inputs along with indices of masked atoms, masked displacements and positions, atom types, and box sizes. Models will then predict the spatial information of the masked atoms and return their predicted displacements $\hat{\mathbf{d}}_i$ relative to other atoms in the same molecule. The to-be minimized objective $\mathcal{L_{\text{masking}}}$ is the negative Cosine Similarity between the prediction $\hat{\mathbf{d}}_i$ and ground true label, $\mathbf{d}_i$:
\begin{equation}
    \hat{d_i} = \phi_\theta(V, \hat{X})
\end{equation}
\begin{equation}
    \mathcal{L_{\text{masking}}} = 1 - \frac{\sum_{i=1}^{n} \hat{d_i} \cdot d_i}{\sqrt{\sum_{i=1}^{n} \hat{d_i} ^2} \cdot \sqrt{\sum_{i=1}^{n} d_i^2}},
\end{equation} 
where $\phi_\theta$ denotes an GNN parametrized by $\theta$. $V$ and $X$ are the respective encoded atom information and Cartesian coordinates of input data samples.

The atom type we choose to mask out is based on the chemical properties of atoms in many molecules that are consist of hydrogens and other heavy atoms. On water system, we empirically discovered that masking Oxygen atoms out causes instability during training and yields degraded results, which could partially be attributed to that recovering Oxygen atom can be too challenging to provide meaningful training signal. Consequently, we decided to mask out a Hydrogen atom from selected molecules and use it as the appropriate pretraining target.

 On the other hand, the remaining atoms would have different impacts on the predictions of masked Hydrogen atoms' coordinates due to their different atomic types. Thus, by reconstructing spatial information of missing atoms, GNNs are expected to learn from restoring the structural and physical information and the inherent structure of water systems. Moreover, the choice of using cosine similarity instead of $L1/L2$ distance to calculate the loss of displacement predictions is due to its scale invariance. It is more reasonable to have the GNNs learn the relative positioning of the atoms instead of the absolute values of the displacements, which may be of small variance under different MD simulation conditions such as changes in Temperature in the data generation process.

\subsection{Pretraining by Denoising} \label{Sec: Denoising Methodology}

Here denoising refers to predicting the artificial noise added to the input atom coordinates in a self-supervised fashion. The noise 
$\xi \in \mathbb{R}^{N\times3}$ is usually sampled from Gaussian $\mathcal{N} (0, \sigma I)$, and such noise is added to atom coordinates as perturbations.

The capacity to predict perturbation exerted over atom coordinates helps GNNs understand the reasonable positioning of molecules and distances relative to each other. As analyzed in several prior works, denoising atomic coordinates is related to learning a pseudo-force field at the perturbed states \cite{zaidi2022pre, xie2022crystal, Wang2023denoise, arts2023two}. For a given molecule with $N$ atoms and its corresponding coordinates $X$, the probability of this conformation ${X}$ is $p(X) \sim \exp{(-E(X))}$ following the Boltzmann distribution, with $E(X)$ being the potential energy at $X$. As shown in \citet{vincent2011}, the network $s_\theta$ the minimizes the following objective:
\begin{equation}
    \frac{1}{2} \mathbb{E}_{q_{\sigma}(\Tilde{X} | X)p(X)}[||s_{\theta}(\Tilde{X}) - \nabla_{\Tilde{X}}(\log q_{\sigma}(\Tilde{X} | X)||_2^2] 
\end{equation}
also satisfies $s_\theta(X)=\nabla_{X}\log q_{\sigma}(X)$, with $\Tilde{X} = X + \xi$ and $\xi \sim \mathcal{N}(0, \sigma I)$. When $\sigma$ is small enough, $s_\theta$ serves as a good estimate of the score function of the true data distribution $\nabla_{X}\log p(X)$. Hence, the denoising pretraining using Gaussian noise with small variance is equivalent to estimating the score function or the negative gradient of the energy.

Given denoising is a popular method to pretrain GNNs when approximating potential energy surface, we also pretrained baseline models using denoising to offer a comparison, as indicated in section \ref{Denoising comparison}.

Specifically, taking noised atom coordinates, atom types, and box sizes as input, GNNs return the predicted noise $\hat{\mathbf{\xi}}$ added to the coordinates. We use the mean squared error (MSE) loss as the objective function of denoising pretraining to capture and minimize the difference between predicted and true noise. The denoising training objective is shown below.
\begin{equation}
    \hat{\xi_i} = \phi_\theta(V, \hat{X})
\end{equation}
\begin{equation}
    \mathcal{L_{\text{denoising}}} =  \|\hat{\mathbf{\xi}} - \xi\|^2 
\end{equation} 
where $\phi_\theta$ denotes an GNN parametrized by $\theta$. $V$ and $X$ are the encoded atom information and noised Cartesian coordinates of input data samples.



\section{Experiment}

\subsection{Experiment Set up on Water Datasets}

\

The whole training pipeline is composed of two stages: pretraining and finetuning. In the pretraining stage, all baseline models are firstly trained on the masked pretraining strategy discussed in Section \ref{Sec: Masking Methodology}. Pretrained model weights are then transferred and finetuned with supervision during the finetuning stage. This transfer aims to leverage the knowledge learned about molecular systems during pretraining. Such knowledge embedded in the GNN parameters is expected to improve the performance of the GNN predicting force and energy labels with a strong prior. The overall architecture is illustrated in Figure\ref{fig:masked architecture}. This training framework is model-agnostic and can easily be applied to different models. 

The water datasets used in this work are generated from classical molecular dynamics (MD) and density functional theory (DFT).
The classical MD simulations are performed using OpenMM \citep{OpenMM} with Tip3p forcefield \citep{TIP3P}. The systems considered are uniform single-atom or single-molecule systems with different simulation box sizes. Periodic Boundary Conditions (PBCs) are set in all directions, and a cut-off distance of 10 angstroms was used. NVT ensemble with a Nosé–Hoover chain thermostat \citep{Nose, Hoover} is used for the simulation. We simulate each configuration of the molecular system to $50 000$ steps with a time step size of $2.0$ femtoseconds and save the state of the system every 50 steps. Each configuration is generated by initializing particles in the system with random positions. A total of $10$ configurations are for generating the data. 
The DFT simulation data of water molecules are obtained from \citet{waterdft-data},
which uses the revised Perdew–Burke–Ernzerhof (RPBE) functional \citep{RPBE} with Van der Waals correction \citep{D3correction} and contains $7241$ structures in total. 

Two water datasets, RPBE and Tip3p, are each split into a train-validation set and a test set with a 90\% and 10\% ratio, and the train-validation set is further split into a train set and a validation set following the same proportion. In our works, we reproduced three previously mentioned Graph Neural Networks, EGNN\cite{satorras2021n}, GNS\cite{kumar2022gns}, and ForceNet\cite{hu-2021-forcenet} as baseline models to evaluate our method's performance. \

GNNs are pretrained on RPBE and Tip3p datasets for 100 and 25 epochs, respectively. The pretrained models are later finetuned over the same dataset. The finetuning epochs are 300 epochs and 50 epochs respectively. The difference between different sets of training epochs mainly comes from the size of the dataset. RPBE has 7241 data samples. Meanwhile, the Tip3p dataset has a total of 10000 data points, each of which comes with 258 water molecules. Thus, GNNs are trained for more epochs to make up for fewer data samples in the RPBE dataset. To better investigate the effectiveness of our masking pretraining strategy, we trained GNNs from scratch without pretraining as baseline models, and we kept the rest of the configurations the same to compare with their pretrained counterparts. Training from scratch epochs are 300 epochs and 50 epochs respectively as a fair comparison. The specific computational cost of pretraining is discussed in Supplemental Information Section 3.

We employ the AdamW optimizer \cite{loshchilov2019decoupled} for both pretraining and finetuning. Learning rates during all stages of training are set to 1e-4 and eventually reduced to 1e-7, with a weight decay rate of 1e-4. We apply the same batch sizes, 16 and 10, for RPBE and Tip3p datasets across training stages and models. Notably, we observed that EGNN is highly sensitive to larger batch sizes, as training would become fairly unstable. Due to our limited computational resources and to create an equal comparison, all models' hidden dimensions are set to 128.
The cutoff radius for water systems is set to 3.4 Angstroms.

\subsection{Experiment Results} \label{sec: Experiment results}

\begin{table}[H]
\centering

\resizebox{\textwidth}{!}{%
\begin{tabular}{cccccccc}

\hline
Datasets & Models & Pretrained & Force RMSE & Energy RMSE & Force Improv & Energy Improv & Masking Ratio \\ \hline
\multirow{6}{*}{RPBE} & EGNN &  & 221.0 (81.3) & 1570.1(660.7) &  &  &  \\
 & EGNN & \checkmark & 90.9(8.3) & 907.6(96.1) & -58.9\% & -42.2\% & 0.25 \\
 & GNS &  & 214.2(74.8) & 1842.2(616.4) &  &  &  \\
 & GNS & \checkmark & 103.2(1.3) & 2216.4(76.9) & -51.8\% & 20.3\% & 0.5 \\
 & ForceNet &  & - & - &  &  &  \\
 & ForceNet & \checkmark & 231.3(5.1) & - & - & - & 0.5 \\ \hline
\multirow{6}{*}{Tip3p} & EGNN &  & 61.5(4.0) & 685.1(55.8) &  &  &  \\
 & EGNN & \checkmark & 39.3(2.0) & 241.9(27.3) & -36.1\% & -64.7\% & 0.9 \\
 & GNS &  & 52.0(0.2) & 638.2(11.6) &  &  &  \\
 & GNS & \checkmark & 49.1(0.3) & 540.4(8.1) & -5.6\% & -15.3\% & 0.5 \\
 & ForceNet &  & - & - &  &  &  \\
 & ForceNet & \checkmark & - & - & - & - & 0.5 \\
 \hline
\end{tabular}
}

\caption{Quantitative analysis of force and energy RMSE on RPBE and Tip3p datasets. Changes in RMSE are shown for pretrained models compared to their trained-from-scratch counterparts, with negative percentages indicating improvements. The units for force and energy RMSE are ($meV/ \si{\angstrom}$) and ($meV$), respectively. All results represent the average of four runs with different random seeds, with the corresponding standard errors reported in parentheses.} 
\label{Tab: Masked experiment results}
\end{table}

Following the described training setup, we conducted experiments to investigate if masked pretraining would help the finetuning objective over energy and force prediction accuracy. All the configurations are trained four times with different random seeds, and we report the averaged results. As shown in Table \ref{Tab: Masked experiment results}, we compared the Root mean square error (RMSE) of predictions of force and energy. The results show that the masked pretraining strategy provides a significant improvement over both objectives across datasets. Most of the RMSE objectives are reduced by pretraining with our proposed masking pretext task. The average improvement of force and energy RMSE for EGNN and GNS is $38.09\%$ and $25.48\%$. Particularly, the pretrained EGNN model gained 47.48\% and 53.45\% of performance improvement on force and energy objectives respectively. Masking pretraining also helps the GNS model reduce $28.71\%$ of force RMSE. This implies that our approach is model-agnostic (equivariant and covariant models) and can be generalized to both force-centric models and energy-centric models.

It is worth noting that the ForceNet model is hardly converging on both datasets without pretraining. Since ForceNet's results are several orders of magnitude higher than other models' results, we did not report out-of-scale results for clarity. However, we observe that finetuning ForceNet becomes more stable after pretraining in comparison to ForceNet trained from scratch. We also observe a similar trend that masked pretraining can effectively stabilize the training of other models, and we argue that masked pretraining could facilitate more stable training and faster convergence. We further demonstrate and elaborate on this point with training loss curves of the ForceNet model in the supplemental information Section 2.

\subsection{Comparison over models trained from scratch with additional epochs}




\begin{figure}[H]
    \centering
    \includegraphics[width=1\linewidth]{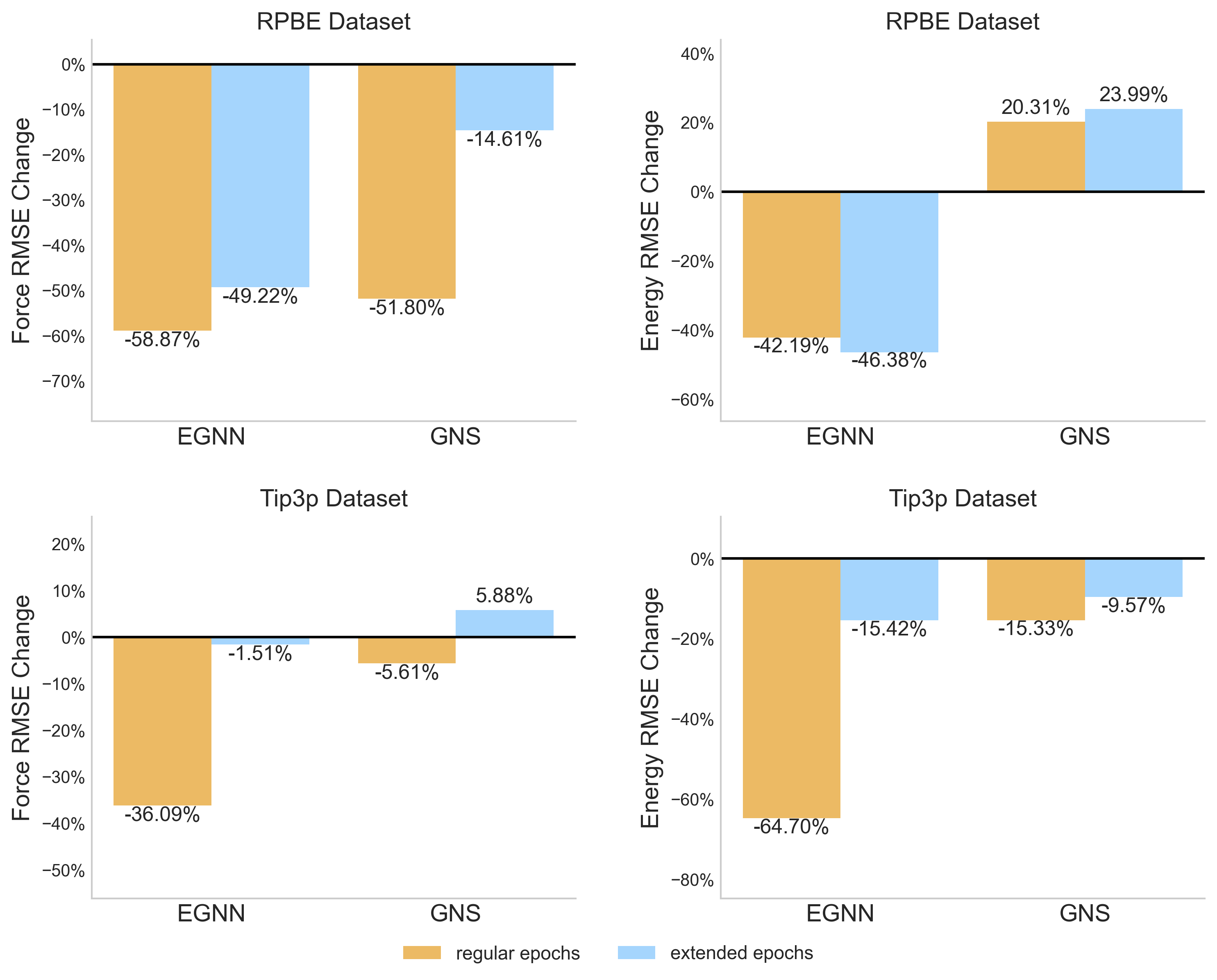}
    \caption{Force and Energy performance comparison between models trained with regular and extended epochs: 300 epochs and 400 epochs on the RPBE dataset, and 50 epochs and 75 epochs on the Tip3p dataset. $\text{Improvement} = 1 - \frac{{RMSE} _\text{(pretrained + regular epochs)}}{{RMSE}_\text{(train from scratch + regular/extended epochs)}}$.
    A negative percentage indicates better performance after pretraining.}
    \label{fig:Full comparison}
\end{figure}



Extended experiments are also conducted to help demonstrate the efficacy and necessity of our two-stage training approach in comparison to simply extending the training duration without pretraining. In contrast to trained from scratch, pretrained models have yielded significant improvements on the RPBE dataset, as shown in figure \ref{fig:Full comparison}. In the first row of figure \ref{fig:Full comparison}, We compared the energy and force RMSE of our pretrained models to the models trained from scratch with either regular 300 or an extended 400 epochs. The results show that our training strategy of 100 epochs of pretraining followed by 300 epochs of finetuning still outperforms the elongated train-from-scratch models.

On the RPBE dataset, the performance gap in force RMSE between EGNN and GNS models trained with and without pretraining sustains, despite the additional finetuning time. It underscores the effectiveness of pretraining over merely increasing the training iterations. On Tip3p, the difference in force RMSE between the pretrained models and the model trained from scratch with extended 75 epochs is reduced, and a similar trend is observed energy-wise. It is worth noting that the pretrained EGNN on the RPBE dataset achieves a higher energy RMSE improvement when compared with EGNN trained from scratch with 400 epochs instead of 300 epochs. We attribute it to the overfitting of the model to this dataset. This observation indicates that for models converging on smaller datasets given certain epochs, starting from a point where the model has learned meaningful molecule featurization (with pretraining), the network is less likely to converge to worse local minima during the final training phase.

Pretraining gives the models a better initialization and facilitates models to extract meaningful representation from the unlabelled data, resulting in more adept and competent models. Therefore, our findings suggest integrating masked pretraining into machine learning potential training pipelines is non-trivial and of great potential.

\subsection{Comparison with Denoising} \label{Denoising comparison}


\begin{figure}[h]
    \centering
    \includegraphics[width=1\linewidth]{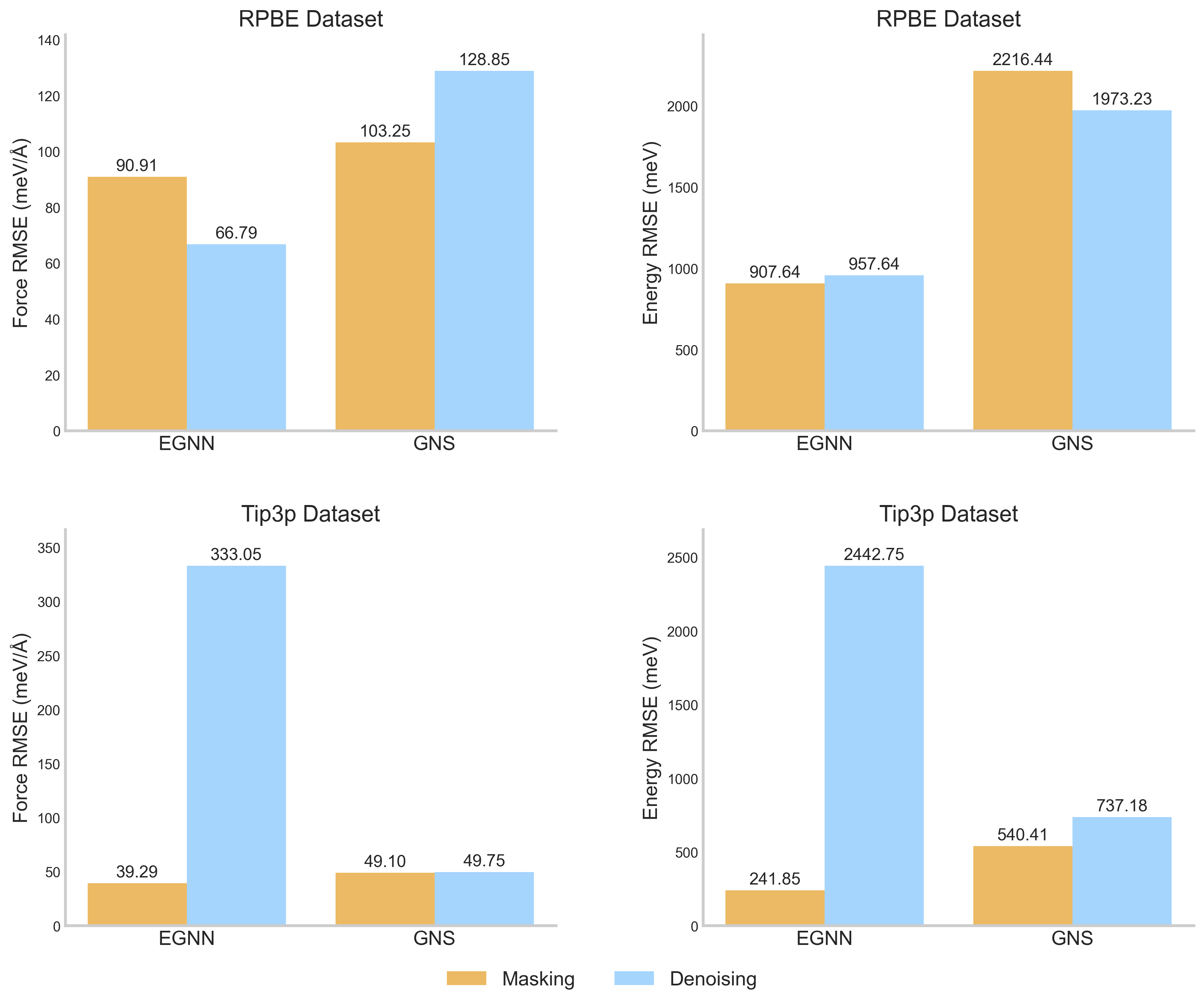}
    \caption{ Force and Energy performance results from models pretrained with masking and denoising tasks on the RPBE and Tip3p datasets} 
    \label{fig:denoise}
\end{figure}
We provide the comparative performance metrics of our proposed masked and denoising methods, which is critical in evaluating the efficacy of different pretraining strategies in predicting molecular properties. Figure \ref{fig:denoise} provided a direct comparison between these two methods across two datasets for force and energy predictions. We held masking pretrained models' results from table \ref{Tab: Masked experiment results} constant. We implemented denoising as described in Sec \ref{Sec: Denoising Methodology}. The mean and standard deviation of the Gaussian Noise added are set to be 0 and 0.5 throughout the experiments.

We observe that in the RPBE dataset, the performance of denoising is similar and competitive to that of our method. We attribute this observation to the effectiveness of the denoising approach in single molecule and attribute label pairs datasets such as ISO17, and ANI-1x as demonstrated in recent research \cite{zaidi2022pre, Wang2023denoise}. Containing on average 27 water molecules in each sample, the number of samples in the RPBE dataset is 77.17\% less than that of the Tip3p dataset. Importantly, denoising has shown difficulties scaling to the Tip3p dataset. EGNN model with masked pretraining has an energy RMSE of about 241.85 $meV$. The performance deteriorates significantly when denoising is applied, with poor convergence and RMSE rising to 2442.75 $meV/ \si{\angstrom}$ by the end of the finetuning stage. Denoising also exhibits degraded performance on the Energy RMSE metric with the GNS model pretrained on the Tip3p dataset. Both EGNN and GNS pretrained using masked pretraining have shown stable convergence during training without suffering from scalability and achieved superior performance.

\subsection{Experiment on Revised MD17 Dataset}\label{sec: MD17} 

We additionally evaluated our masked pretraining method with EGNN on the Revised MD17 dataset \cite{christensen2020role}, which is a refined version of the MD17 dataset \cite{Sch_tt_2017, chmiela2018towards}. It contains molecular dynamics (MD) trajectories of 10 types of small organic molecules at DFT accuracy, each comes with 100,000 samples (except for Azobenzene which has 99,888 samples). \

Following the setup in  \citet{schutt2017schnet}, the training sample size is 50,000, with 1,000 samples for evaluation, and the rest for testing. For Pretraining, we combine all the training samples for each molecule as the pretraining dataset and further finetune the models on the individual molecules. Thus, the overall pretraining dataset is of size 500,000 for 10 molecules. 
After pretraining, the pretrained model checkpoint is subsequently finetuned on each molecule individually with the same data splits to ensure no data leakage. 


We adopt the number of pretraining epochs to 15 and finetuning epochs to 200 in addition to a batch size of 32. Following prior works, we use Adam Optimizer with a learning rate of 1e-4 and a minimum learning rate of 5e-5. Prior works \cite{TorchMD-2021-JCTC, liao2023equiformer} regard force and energy weights as tunable hyperparameters during training. In this work, we set them to 40 and 15 respectively, with which we observed reasonable convergence of EGNN on the Revised MD17 dataset. 

We pretrained EGNN models using three masking ratios (0.25, 0.5, 0.75) and then fine-tuned them on the Uracil molecule as validation. Although all three models showed reasonable performance improvements, we transferred the model weights pretrained with a masking ratio of 0.5 and finetuned each molecule in the dataset on top of it for its best performance among all.

\begin{table}[ht]
\centering
\begin{tabular}{llcccccc}
\hline
Molecule & Pretrain & Force MAE & Energy MAE & \multicolumn{2}{c}{Change (\%)} \\ 
 &  &  &    & Force & Energy \\ 
\hline
Revised Uracile &  & 22.0 & 11.8 & & \\
 & \checkmark & 19.0 & 10.7 & -13.6\% & -9.3\% \\
\hline
Revised Benzene &  & 14.6 & 5.4 & & \\
 & \checkmark & 7.7 & 6.2 & -47.3\% & 14.8\% \\
\hline
Revised Naphthalene &  & 14.6 & 30.3 & & \\
 & \checkmark & 18.5 & 14.7 & 26.7\% & -51.5\% \\
\hline
Revised Aspirin &  & 46.7 & 25.2 & & \\
 & \checkmark & 29.5 & 16.0 & -36.8\% & -36.5\% \\
\hline
Revised Malonaldehyde &  & 32.9 & 14.4 & & \\
 & \checkmark & 11.7 & 8.3 & -64.4\% & -42.4\% \\
\hline
Revised Ethanol &  & 16.8 & 8.8 & & \\
 & \checkmark & 10.7 & 4.7 & -36.3\% & -46.6\% \\
\hline
Revised Toluene &  & 17.9 & 13.7 & & \\
 & \checkmark & 12.1 & 10.2 & -32.4\% & -25.5\% \\
\hline
Revised Salicylic Acid &  & 27.5 & 25.9 & & \\
 & \checkmark & 17.6 & 13.2 & -36.0\% & -49.0\% \\
\hline
\end{tabular}
\caption{Force MAE and Energy MAE performance change results for EGNN with or without masked pretraining on the Revised MD17 dataset. The units for force and energy MAE are  ($meV/ \si{\angstrom}$) and ($meV$).}
\label{table: MD17 improvement}
\end{table}

We present the results for Force and Energy mean absolute error (MAE) in Table \ref{table: MD17 improvement}, showing an average reduction of 30.02\% and 30.58\%, respectively. These findings suggest that our method can effectively generalize to small organic molecules with diverse types of atoms that are heavier than the Hydrogen atom. A major advantage of our proposed strategy is that it can be trained in a diverse set of unlabelled configurations, after which the pretrained model can be transferred to different types of molecules. Additionally, initializing from the weights that have already learned general patterns from extensive data, the convergence of the pretrained model is also faster, allowing the model to quickly adapt to different molecular structures. Most importantly, this approach ensures consistent and reliable performance across different molecules, which is vital for robust and accurate potential energy approximation.

\section{Conclusion}

We present a new method of masked pretraining for improving GNN's performance on forcefield learning. We demonstrate that our method could achieve better accuracy on both force and energy prediction. Moreover, it also facilitates more stable training and exhibits better convergence in comparison to models without pretraining or pretrained with other pretext tasks such as denoising. These results indicate the effectiveness of our method for further utilization of GNNs to facilitate the development of neural potentials. Despite our method being currently limited to molecules that contain Hydrogen atoms, we believe that exploring masking other atoms or masking a combination of different types of atoms and scaling our method with larger and more advanced GNNs, are promising directions for future work.

\section{Supplementary Material}
Supplementary material includes the following sections:
\begin{itemize}
  \item Section 1: Ablation study of the choice of masking ratios;
  \item Section 2: Further analysis of the ForceNet\cite{hu-2021-forcenet} results mentioned in Sec \ref{sec: Experiment results};
  \item Section 3: Computational cost of masked pretraining;
\end{itemize}

\section{Data Availability Statements}
The data and code that support the findings of this project can be found at: \url{https://github.com/StevenZhang904/Masked_Pretrain.git}.

\bibliography{ref}

\end{document}


\section{Choice of Masking Ratios} \label{Sensitivity}
Throughout the experiments, we found that many factors directly influence the final performance, and among all, the choice of masking ratio is rather important to analyze. The masking ratio relates to the difficulty of pretraining tasks during pretraining, thus determining how much pretrained models can learn about the underlying dynamics of molecule systems. A higher masking ratio for pretraining leaves the model with the harder task of recovering spatial information of more missing atoms. We conducted additional experiments on training EGNN with three different masking ratios and compared the validation RMSE curve with the trained-from-scratch ones on both datasets. 

\begin{figure}[H]
    \centering
    \includegraphics[width=1\linewidth]{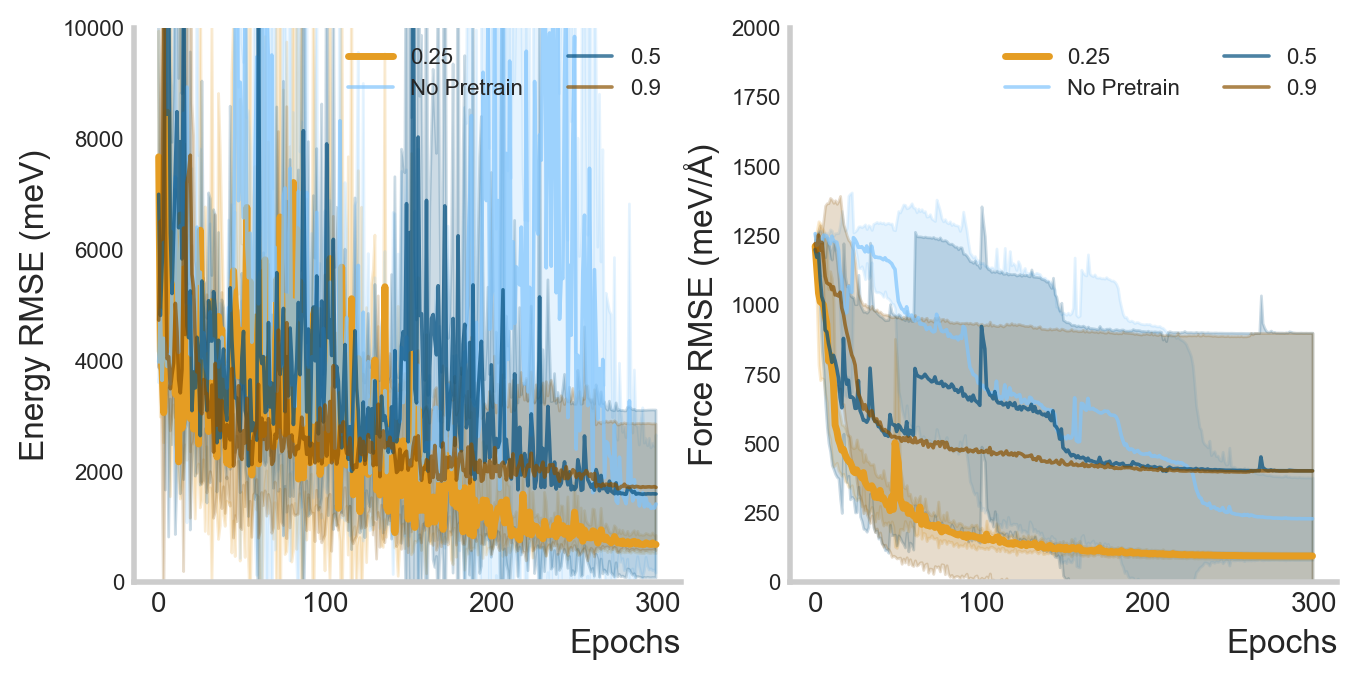}
    \caption{Validation RMSE curve of finetuning EGNN model, which is pretrained with different masking ratios, on the RPBE dataset. The results of the masking ratio of 0.25 are selected and reported in Table 1 of the main manuscript, shown in the above figures as the bold yellow line.} 
    \label{fig:EGNN masking ratio comparison RPBE}
\end{figure}

\begin{figure}[H]
    \centering
    \includegraphics[width=1\linewidth]{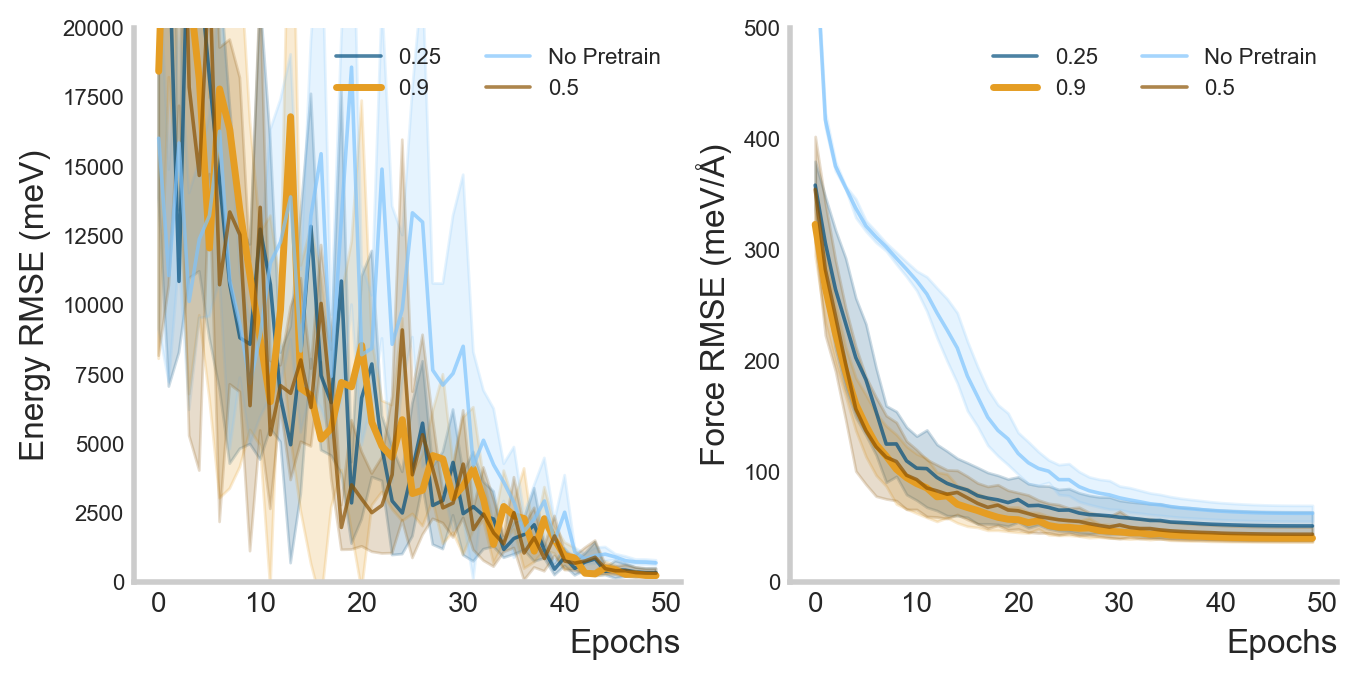}
    \caption{Validation RMSE curve of finetuning EGNN model, which is pretrained with different masking ratios, on the Tip3p dataset. The results of the masking ratio of 0.9 are selected and reported in Table 1 of the main manuscript, shown in the above figures as the bold yellow line.} 
    \label{fig:EGNN masking ratio comparison Tip3p}
\end{figure}

In general, a higher masking ratio on a more diverse and challenging dataset like the RPBE dataset would make the model harder to learn good prior from it with limited epochs of pretraining and lead to degraded results, as shown in figure \ref{fig:EGNN masking ratio comparison RPBE}. A relatively lower masking ratio of 0.25 achieves better convergence throughout the training duration. Meanwhile, when trained on the Tip3p dataset as shown in figure \ref{fig:EGNN masking ratio comparison Tip3p} where the configuration is less diverse, the model can restore the information with 90\% of water molecules having masked hydrogen atoms. In addition, when trained on the Tip3p dataset, all three choices of masking ratio could achieve better performance, indicating the robustness of our pretraining strategy. In practice, we suggest selecting a masking ratio based on the size of the dataset to avoid negative transfer and justify the trade-off between extra training time required by the pretraining stage and performance gain. 

\section{ForceNet Convergence Analysis} \label{ForceNet convergence}

\begin{figure}[h]
    \centering
  \begin{subfigure}[b]{0.95\textwidth}
    \includegraphics[width=\textwidth]{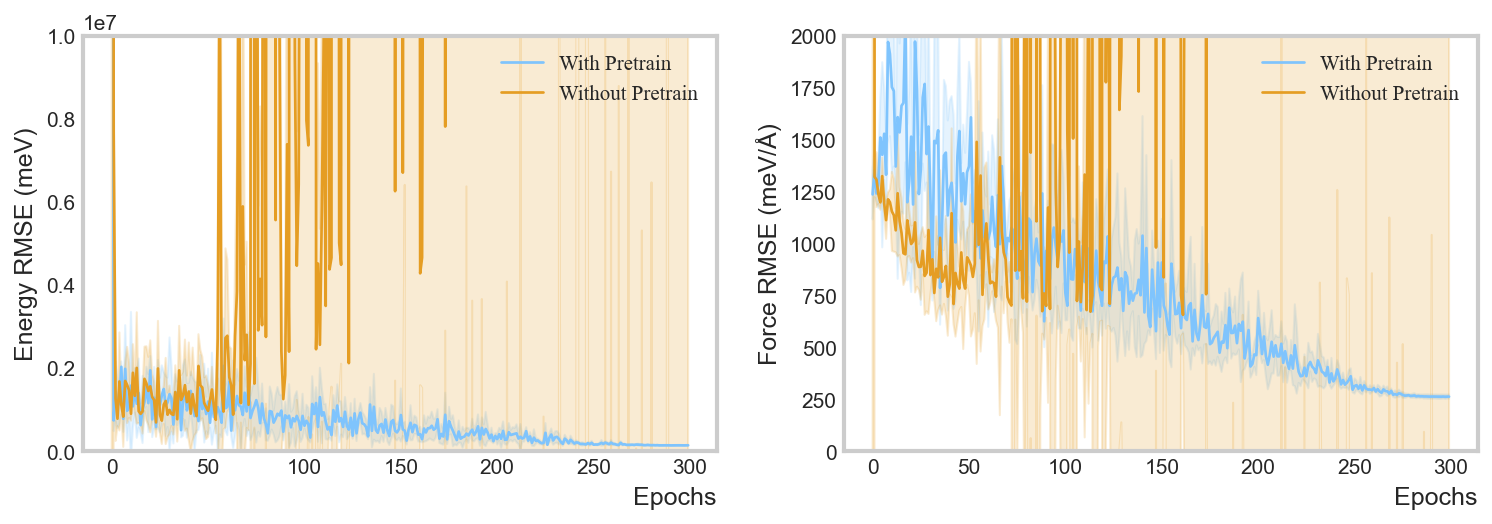}
    \caption{Validation RMSE curve of ForceNet finetuned on RPBE dataset}
    \label{fig:1}
  \end{subfigure}
 \hfill
  \begin{subfigure}[b]{0.95\textwidth}
    \includegraphics[width=\textwidth]{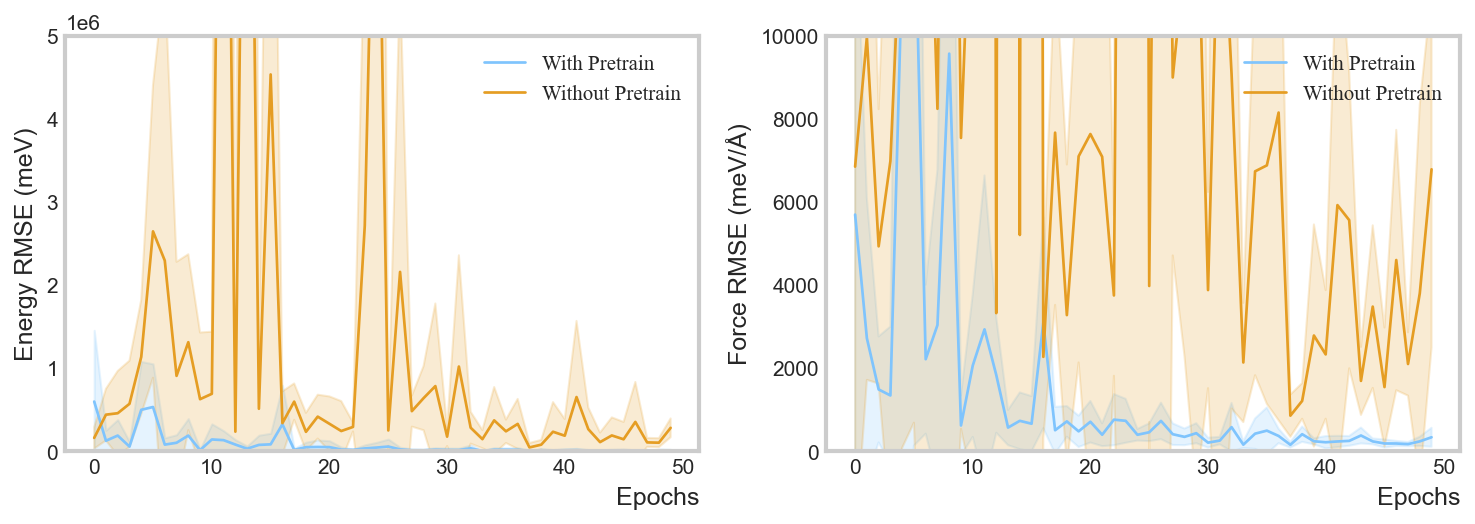}
    \caption{Validation RMSE curve of ForceNet finetuned on Tip3p dataset}
    \label{fig:2}
  \end{subfigure}
  \caption{Finetuning validation RMSE curve of ForceNet model trained with and without pretraining  }
  \label{fig: ForceNet_pretrained}
\end{figure}

Figure \ref{fig: ForceNet_pretrained} illustrates that masking pretraining can effectively stabilize the training process. Pretrained ForceNet models display smoother validation RMSE curves across both datasets, whereas models trained from scratch exhibit more erratic fluctuations and show limited signs of convergence. In particular, the ForceNet model trained from scratch on the RPBE dataset experiences gradient explosions in both energy and force objectives. Although the RPBE dataset contains fewer data points, it features greater diversity compared to the Tip3p dataset sampled from empirical molecular dynamics. This diversity is hypothesized to be the primary cause of ForceNet's instability when using the RPBE dataset. Generally, smaller datasets often require additional training epochs to develop effective representations and are prone to experiencing gradient explosions, leading to oscillations around local minima.

\section{Pretraining Computational Cost}\label{sec: Computational Cost}\

The additional computational cost is a critical component when considering adopting our proposed pretraining method. We record the pretraining time duration for each model and dataset in Table 1 of the main manuscript. All experiments are conducted on one NVIDIA RTX 4090 with 24GB VRAM running on WSL2, which may cause computational overhead and slightly slow down training.\

In general, the pretraining cost is smaller than training on labeled datasets, and meanwhile, it provides a notable performance improvement. For most models and datasets, the pretraining can be finished in 1 GPU hour. The exception is EGNN pretrained on the rMD17 dataset. On rMD17 datasets, we are combining all the training samples for different species of the molecule as the pretraining dataset and further finetuning the models on the individual molecules. Therefore our pretrained model could be reused for all molecules without the need for individual pretraining for each one of them. \

\begin{table}[htbp]
\centering
\resizebox{\textwidth}{!}{%
\begin{tabular}{@{}cccc@{}}
\toprule
\textbf{Dataset} & \textbf{Model} & \textbf{Total Pretraining Time (Hours)} & \textbf{Finetuning from Scratch Time (Hours)} \\ \midrule
\multirow{4}{*}{Water - RPBE} & EGNN & 0.57 & 2.36 \\
 & ForceNet & 0.51 & 2.14 \\
 & GNS & 0.64 & 2.13 \\
  \midrule
\multirow{3}{*}{Water - Tip3p} & EGNN & 0.76 & 2.72 \\
 & ForceNet & 0.7 & 1.75 \\
 & GNS & 0.6 & 1.83 \\ \midrule
Revised MD17 & EGNN & 1.75 & Varies for each molecule \\ \bottomrule
\end{tabular}%
}
\caption{Model Training and Finetuning Details. All models trained on RPBE and Tip3p datasets were trained with a sample size of 5792 and 8000 respectively, with 100 and 25 pretrain epochs, and 300 and 50 finetune epochs correspondingly. The Revised MD17 model was trained with a sample size of 500,000, 15 pretrain epochs, and 200 finetune epochs.}
\label{table: Computational cost}
\end{table}